\documentclass{article} 
\usepackage{iclr2027_conference,times}

\usepackage{amsmath,amsfonts,bm}

\def\eqref#1{equation~\ref{#1}}

\def\1{\bm{1}}

\DeclareMathAlphabet{\mathsfit}{\encodingdefault}{\sfdefault}{m}{sl}
\SetMathAlphabet{\mathsfit}{bold}{\encodingdefault}{\sfdefault}{bx}{n}

\usepackage{hyperref}
\usepackage{url}
\usepackage{booktabs}
\usepackage{array}
\usepackage{enumitem}
\usepackage{subcaption}
\usepackage{wrapfig}
\usepackage{longtable}
\usepackage{float}
\usepackage{graphicx}
\usepackage{tcolorbox}
\tcbuselibrary{breakable,skins}
\usepackage{newunicodechar}
\usepackage{amssymb}
\newunicodechar{【}{[}
\newunicodechar{】}{]}

\newcolumntype{L}[1]{>{\raggedright\arraybackslash}p{#1}}
\newcolumntype{C}[1]{>{\centering\arraybackslash}p{#1}}

\newcommand{\benchname}{RITE}
\newcommand{\benchfull}{Resilience to Incorrect Tool Evidence}
\newcommand{\ws}[1]{\textcolor[rgb]{1.0, 0.4, 0.4}{#1}}

\iclrfinalcopy
\title{RITE: Evaluating End-to-End Resilience to\\ Silent Tool Errors in E-Commerce Agentic Systems}

\author{
Wenshuo Zhang$^{1}$, Yuquan Lin$^{1}$, Yushi Sun$^{1}$, Yishi Lin$^{2}$, Rui Sheng$^{1}$,
 Peng He$^{2}$, Han Gao$^{2}$, \\ \bfseries  Jikai Wang$^{2,\dag}$, Yifeng Rong$^{2,\dag}$\\
$^{1}$The Hong Kong University of Science and Technology\quad
$^{2}$Tencent\\
\texttt{wzhangeb@connect.ust.hk, \{crazyrong, teddyjkwang\}@tencent.com}\\
\vspace{1mm}$^{\dag}$Co-corresponding authors.
}

\newcommand{\task}{Silent-error Resilience Task}

\begin{document}

\maketitle

\begin{abstract}

Tool-using LLM agents increasingly answer e-commerce queries through multi-step tool interactions. The expected output of a tool call can shift between agent development and execution, allowing a plausible but incorrect value to be returned without an explicit failure signal.
These silent errors may propagate through subsequent decisions into final answers. 
To study agents' resilience across this end-to-end process, we formulate \task{}, an interactive task that evaluates whether they complete the original query after a silent error corrupts an intermediate state.
We introduce \benchname{}, a benchmark built from human-authored industrial workflows and production-log-derived inputs. To enable controlled and reproducible evaluation, we use a semantic error taxonomy to inject errors into fixed workflow prefixes, freeze recorded tool behavior, and let agents continue freely.
It contains 167 matched units across nine e-commerce workflows, yielding 501 silent-error, correct-value, and explicit-failure instances. 
Across 10 recent LLMs, silent errors roughly halve task success and are consistently harder than explicit failures. Recognition, use of an available verification path, and integration of the resulting evidence form successive bottlenecks.
We also propose a light State Checker as an intervention to test whether resilience can be improved.

\end{abstract}

\section{Introduction}
\label{sec:intro}

Agentic LLM systems increasingly complete e-commerce tasks through multi-step workflows that invoke multiple external tools and pass their outputs from one step to the next.
In industrial deployments, a tool's runtime output may differ from what the agent was built to expect because underlying business data, tool implementations, or deployed versions can change between development and execution.
This build-time--runtime mismatch can remain hidden: a call may complete and conform to the expected API schema while returning a plausible value that is incorrect for the current task.
We refer to such plausible-but-wrong intermediate values without an explicit failure signal as \emph{silent errors}. 
For example, suppose a user mainly watches local football videos, one of which mentions blockchain-based fan tokens. During agent development, the profile tool returns ``local sports''; after an update that overweights ``blockchain,'' it returns ``blockchain'' at execution. The new label remains plausible and schema-valid, so no failure is signaled.
The absence of an explicit failure signal makes silent errors particularly severe: an agent may have no immediate basis for detecting that the value is incorrect. It may therefore propagate the corrupted value through subsequent tool calls, allowing a single intermediate error to directly alter the final user-facing output.

Existing benchmarks evaluate tool use in nominal settings \citep{qin2024toolllm,trivedi-etal-2024-appworld,ma2024agentboardanalyticalevaluationboard}, recovery from explicit or implicit faults injected while agents run \citep{zhu2026toolsfailbenchmarkingdynamic,tian2026functioncallingbenchmarkingtoolusing,chen2026retryswitchabstainlearning}, silent-error detection \citep{sun-etal-2024-tools,mazumder2026agentcheckreproduceintervenemitigateworkbenchllm}, or how corrupted intermediate values affect later outputs \citep{gurram2026auditingautomatedevaluationerror}.
In these fault-based settings, each agent runs freely and meets a fault only if and when its own trajectory reaches the perturbed call, so different agents face silent errors at different points and with different histories. They therefore cannot compare different agents on the same silent error at the same decision point, nor pair that error with correct-value and explicit-failure counterparts at that point.

To address this gap, we formulate \task{}, in which a plausible but incorrect tool result enters an agent's execution history without an explicit failure signal.
Given the query and toolbox, the agent operates in an agentic loop, repeatedly reasoning, calling tools that may return silent errors, and incorporating their outputs until it produces a final answer.
The task asks whether the agent can recognize the error, reacquire reliable evidence, and complete the original query instead of propagating the error into its final answer.

To evaluate this task under controlled conditions, we introduce \benchname{} (\benchfull{}).
Each human-authored workflow specifies a pre-orchestrated agentic process, and its execution log records tool calls and observations in the same format as an agent trajectory. We use this log as a fixed trajectory, allowing us to control where a silent error is introduced.
We select a semantically important slot in the log and inject a silent error at that point. 
The trajectory up to and including this slot is provided as the agent's prefix, while the workflow's tools and their recorded input--output behavior are frozen to form the evaluation world, so that recorded calls return identical outputs across models and conditions. 
The agent then continues in an agentic loop, allowing us to compare its behavior while controlling for error type, slot, and environment.

Following this protocol, we first construct a semantic taxonomy covering errors in object or category selection and in attributes or values. 
We sample inputs from the production logs of one internal platform and use them to run workflows deployed on a separate internal platform.
From the resulting traces, we select task-relevant checkpoints and instantiate taxonomy-guided plausible but incorrect values.
Each resulting unit contains matched prefixes for the silent-error, correct-value, and explicit-failure conditions, all paired with the same evaluation world. \benchname{} contains 167 matched units from nine e-commerce workflows, yielding 501 instances.

We evaluate 10 recent closed and open-weight LLMs on \benchname{}. Across all models, silent errors reduce task success to roughly half of correct-condition performance and are consistently harder to handle than matched explicit failures. Our analysis traces this weakness through a three-stage chain: agents often fail to \emph{recognize} that an inherited value may be wrong; recognition does not reliably lead them to \emph{refetch} its source; and refetching does not guarantee that the corrected value is carried through to final \emph{recovery}. We further find that most benchmark corruptions would change the downstream e-commerce task output, with damage varying substantially by error type and workflow position. Finally, a lightweight State Checker makes agents notice and refetch more often but yields only limited improvement in final task success. Making agents look is therefore easier than making them repair what they find. In summary, our contributions are:
\begin{itemize}
    \item \textbf{Task formulation.} We formulate \task{} as an interactive agentic task for studying resilience to silent errors throughout the query-to-answer process. It evaluates whether an agent can recognize an unreliable tool result, reacquire trustworthy evidence, and still produce a correct final answer rather than propagate the error downstream.
    \item \textbf{Benchmark construction.} We introduce \benchname{}, which replays a human-authored workflow execution up to a selected point, returns a correct, silently incorrect, or explicitly failed result there, and lets the agent continue freely. Using inputs sampled from one internal platform's production logs, workflows deployed on another, and a semantic error taxonomy, we construct 167 matched units across nine e-commerce workflows, comprising 501 silent-error, correct-value, and explicit-failure instances.
    \item \textbf{Empirical analysis.} Across 10 LLMs, silent errors roughly halve task success. We identify recognition, producer refetch, and recovery as successive failure points, characterize effects by error type and position, and propose a lightweight State Checker as an intervention to test whether resilience can be improved.
\end{itemize}

\section{Related Work}
\label{sec:related}

\paragraph{Tool-use and agent evaluation.}
Agent evaluation spans three levels. 
At the tool-call level, MetaTool \citep{huang2024metatool} asks whether a tool is needed and which one to select, while ToolBench, BFCL, and API-Bank \citep{qin2024toolllm,patil2025the,li-etal-2023-api} evaluate call construction and extend to multi-tool, dependent, and stateful multi-turn interactions; ToolACE \citep{liu2025toolace} provides synthesized data for improving these abilities. 
At the task-outcome level, AppWorld, GAIA, and WebArena \citep{trivedi-etal-2024-appworld,ICLR2024_25ae35b5,zhou2024webarena} judge a complete rollout by its final answer or by the state it leaves in the environment, while allowing different valid action paths. 
At the process level, AgentBoard \citep{ma2024agentboardanalyticalevaluationboard} tracks task progress, WorFBench \citep{qiao2025benchmarkingagenticworkflowgeneration} evaluates generated workflow structure, ToolScan \citep{kokane2025toolscan} diagnoses call-level errors, and process-mining methods \citep{b9ce1ceaafbe4a20bc39b8b6026ea9be} compare executed and expected flows. 
Across these levels, the benchmarks above vary task difficulty and agent behavior but do not systematically manipulate the semantic correctness of intermediate tool values. 
We instead vary the intermediate business value and ask whether an agent can still complete the task correctly when that value is wrong. To keep the test controlled and reproducible, we provide the recorded workflow execution up to the selected checkpoint as the agent's initial history, then let it continue freely.

\paragraph{Tool-failure recovery, silent-error detection, and error propagation.}
Work on unreliable tools can be separated by the stage of the error-handling chain that it evaluates.
One line studies recovery from tool failures: agents resume from the environment left by a failed attempt \citep{tan2025recoverybench}, rewind to an earlier checkpoint \citep{zhuang2026agentrewindrecoverableexecutionlonghorizon}, recover from explicit or implicit faults injected into tool responses while the agent runs \citep{zhu2026toolsfailbenchmarkingdynamic,tian2026functioncallingbenchmarkingtoolusing,chen2026retryswitchabstainlearning}, or diagnose failed trajectories and characterize recovery cost \citep{vuddanti2026recoverabilitylawerrmeasure,zhu2025llmagentsfaillearn,zhang2026agentracer}.
Resumption settings start from uncontrolled natural failures, while injection settings expose an agent to a fault only if and when its own trajectory reaches the perturbed call.
A second line studies silent-error detection, asking whether agents identify faulty outputs that trigger no exception \citep{sun-etal-2024-tools} or silent data-quality faults introduced during controlled replay \citep{mazumder2026agentcheckreproduceintervenemitigateworkbenchllm}.
In both, the faulty response is delivered when the agent's own run issues the corresponding call.
A third line studies error propagation, tracing a schema-valid but semantically wrong argument to the final answer \citep{gurram2026auditingautomatedevaluationerror} and measuring how perturbed requests or tool responses alter later calls and task outcomes \citep{xiong-etal-2025-butterfly,wang2026agentnoisebenchbenchmarkingrobustnesstoolusing}.
To our knowledge, \benchname{} is the first benchmark to evaluate business-semantic silent errors at controlled decision points: every agent encounters the same plausible but incorrect value after the same recorded execution history, the same point is also instantiated with the correct value and with an explicit failure, and the agent then continues freely to a final answer.

\paragraph{E-commerce and business-domain agents.}
Work in this domain spans three forms of capability. 
Domain models test whether a model understands products and users well enough to answer questions \citep{peng2024ecellmgeneralizinglargelanguage}, while recommender agents use that understanding to retrieve, rank, and recommend products \citep{gao2023chatrecinteractiveexplainablellmsaugmented,huang2024recommenderaiagentintegrating}. 
Shopping benchmarks place agents in interactive stores, where they must search, compare, and purchase while satisfying constraints on products, vouchers, budgets, and sellers \citep{yao2023webshopscalablerealworldweb,wang2026shoppingbenchrealworldintentgroundedshopping}. 
Business-service benchmarks require agents to interact with users over multiple turns, operate on business records through tools, and comply with domain policies \citep{yao2025taubench,wang-etal-2025-ecom}, while CRMArena-Pro additionally tests confidentiality in enterprise workflows \citep{huang2026crmarenapro}. 
These benchmarks do not systematically test what happens when a task-critical business value is silently corrupted. 
Grounded in production execution logs and human-authored industrial e-commerce workflows, our benchmark instead examines this reliability risk involving tools developed and maintained by multiple parties: a call may complete while an intermediate business value is silently wrong.

\section{Problem Formulation}
\label{sec:formulation}
We study whether an agent can remain reliable when a tool silently returns an incorrect result and still complete the original task. We first formulate this end-to-end resilience as \task{}, then define how \benchname{} evaluates it in this section.

\paragraph{\task.}
In a standard agentic task, an agent $A$ receives a task query $q$ and repeatedly selects and executes tool calls from a tool set $\mathcal{T}$ provided by an environment $E$. 
The interaction begins without a preloaded execution history, i.e., $h_0^{\mathrm{std}}=\varnothing$. At each step $t$, the agent observes the current history $h_t$ and selects an action $a_t$. The environment executes the action and returns an observation $o_t$, which is appended to the history:
\begin{equation}
    a_t=A(q,h_t,\mathcal{T}), \qquad
    o_t=E(a_t), \qquad
    h_{t+1}=h_t\oplus(a_t,o_t).
    \label{eq:agent-loop}
\end{equation}
The standard action set is $\mathcal{A}_0=\{\text{call},\text{reason},\text{stop}\}$. The interaction ends when the agent stops or reaches the step budget, producing a final answer.

\begin{wrapfigure}[11]{r}{0.55\linewidth}
\vspace{-1pt}
\centering
\includegraphics[width=\linewidth]{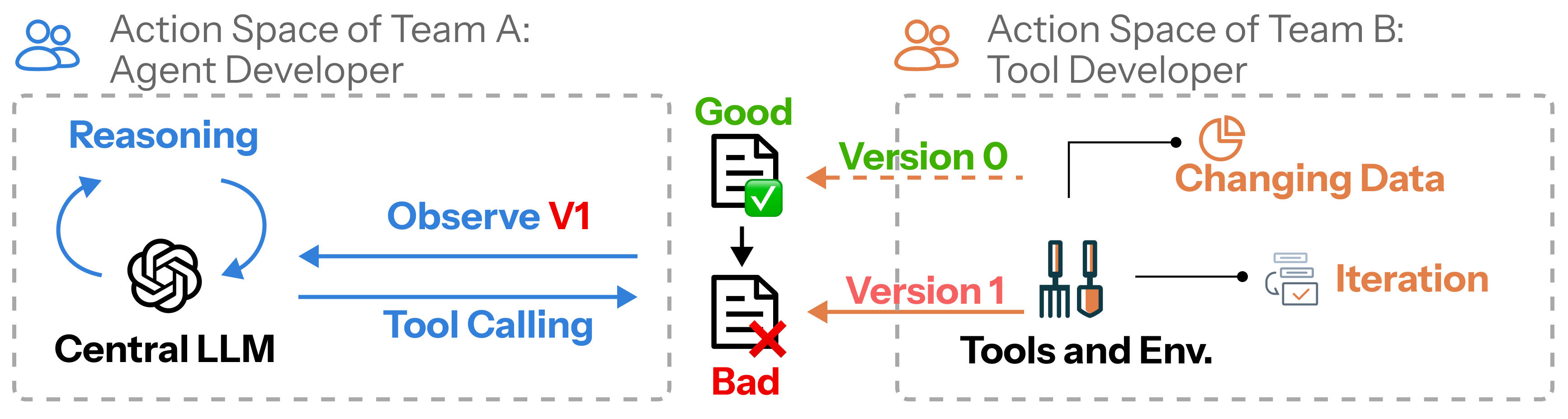}
\caption{Silent errors from build-time--runtime mismatch. Tool-side data or implementation changes can make a runtime result diverge from what the agent was built to expect while signaling no failure.}
\label{fig:silent-error-setting}
\vspace{-4pt}
\end{wrapfigure}

As illustrated in Figure~\ref{fig:silent-error-setting}, the standard loop does not guarantee that a returned observation is semantically correct. In deployed systems, changes to business data, tool implementations, or deployed versions can cause runtime results to diverge from the expectations under which the agent was built. 
A call may still complete and conform to its API contract while returning a result that is incorrect for $q$. 
At a tool-call step $j$, let $o_j^\star$ denote the task-correct result and $\tilde{o}_j$ the result actually returned by the tool. 
A \emph{silent error} occurs when $\tilde{o}_j$ differs from $o_j^\star$ in task-relevant meaning but remains plausible, schema-valid, and free of an explicit failure signal. 
Because the ordinary update rule still appends $(a_j,\tilde{o}_j)$ to the history, subsequent actions condition on unreliable evidence and may propagate it into the final answer.

The \task{} measures whether the agent can still complete $q$ after such a result has entered its execution history. 
Process measures track whether the agent recognizes the unreliable result, reacquires trustworthy evidence, and integrates that evidence into a correct final answer.
The agent retains $\mathcal{A}_0$ and may reason from the current state, revisit or reissue an earlier tool call, seek corroborating evidence, or pursue another continuation. 

\paragraph{Benchmark evaluation.}
For each benchmark unit $i$, \benchname{} constructs three matched instances $D_i^k=(q_i,p_i^k,c_i)$ for $k\in\{\mathrm{correct},\mathrm{implicit},\mathrm{explicit}\}$. The instances share the query $q_i$, result constraint $c_i$, and continuation environment $E_i$, keeping the user task, required result, and available recovery paths fixed. They differ only in the affected evidence in $p_i^k$: the correct condition provides the recorded value as a baseline, the implicit condition provides a silently incorrect value, and the explicit condition provides an overt failure to expose the effect of a visible failure signal.

For each condition, \benchname{} initializes the agent with $h_{0,i}^k=p_i^k$ and provides $q_i$ as the task query. Starting from the matched checkpoint controls variation in the preceding execution, while free continuation in $E_i$ with the standard action set $\mathcal{A}_0$ lets the agent choose its own recovery path. The run ends when the agent stops or reaches the step budget.
Let $\hat{r}_i^k$ denote the agent-generated continuation, which ends with a final answer $y_i^k$; the complete trace is
$r_i^k=p_i^k\oplus\hat{r}_i^k$.

We use three evaluation measures. \emph{Noticed} records whether the agent explicitly identifies a content or result anomaly and takes a related remedial action. Result Success Rate (RSR) records whether $y_i^k$ resolves $q_i$, capturing end-to-end task success. Retention, the ratio of implicit-condition RSR to correct-condition RSR, reports how much baseline performance is preserved under silent errors.

\section{Benchmark Construction and Evaluation Environment}
\label{sec:data}

Each human-authored workflow specifies a pre-orchestrated agentic process. Its execution produces a controlled trace of tool calls and observations in the same format as an agent rollout. 
We use this trace to construct matched prefixes that share the preceding history and differ only in selected intermediate evidence; all prefixes are paired with the same continuation environment.
This section first describes the data sources and silent-error taxonomy, then explains how each clean trace yields paired benchmark instances $D_i^k$ and an evaluation environment $E_i$, and reports its distribution.

\subsection{Data Sources and Silent-Error Taxonomy}
\label{sec:data:source}

\paragraph{Data sources.}
Benchmark construction begins with two source collections.
The first is the workflow collection $\mathcal{W}$, drawn from human-authored e-commerce workflows already deployed and running on an internal industrial workflow platform.
We screen these existing workflows for structural validity and task diversity, yielding nine golden workflows with explicit control flow and deterministic tool dependencies.
The second source is the input collection $\mathcal{X}$ from a separate internal e-commerce data platform. Candidate production-log records are mapped to workflow inputs and selected only when their executions yield complete traces with task-relevant intervention points; the corresponding traces are retained. Appendix~\ref{app:data-sources} defines the workflow- and log-selection criteria.

\paragraph{Taxonomy definition.}
We constructed the silent-error taxonomy through a targeted, literature-grounded synthesis of adjacent error and failure taxonomies. We selected studies whose source categories could clarify the boundary of a silent state corruption or be operationalized as schema-valid, plausible-but-wrong workflow values. For each relevant source category, two authors independently assessed its observable error signal, workflow locus, and suitability for conversion into a single-slot implicit perturbation. They then reconciled overlapping categories and clarified the resulting boundaries.
The final taxonomy contains seven leaf types organized into two semantic groups. 
Object or category selection errors cover substitution with a same-level category (category sibling drift), an ancestor or descendant (category granularity mismatch), or a similar but incorrect real entity (entity near-duplicate). 
Attribute or value errors cover changes to an entity's attribute (profile attribute drift), movement to a neighboring valid numeric range (numeric bucket shift), superficially similar text with a different business meaning (semantic synonym drift), and reuse of a formerly valid but outdated value (temporal staleness). 
The benchmark instantiates every leaf type except temporal staleness. Although a stale value can be schema-valid and lead to an incorrect answer, recognizing that it is stale requires prior or time-indexed evidence that the agent does not observe. Appendix~\ref{app:taxonomy-evidence} provides the complete definitions, category boundaries, and literature grounding.

\subsection{Benchmark Infrastructure Construction}
\label{sec:data:pipeline}
\label{sec:data:infrastructure}

\begin{figure}[t]
    \centering
    \includegraphics[width=\linewidth]{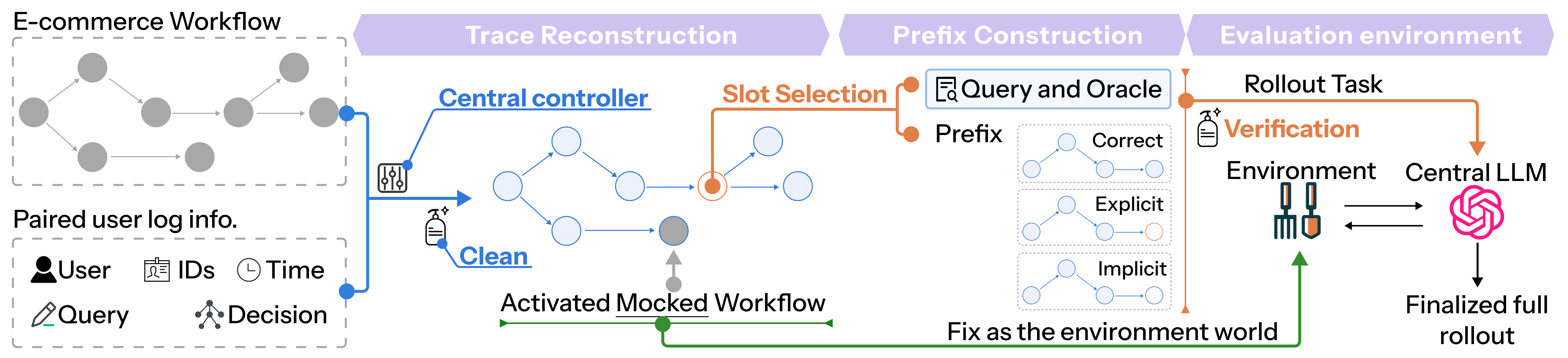}
    \caption{\benchname{} benchmark construction and rollout pipeline. A workflow $W_i$ and paired log-derived input $x_i$ are used to activate a side-effect-free workflow replica and reconstruct a clean trace $\tau_i$. A task-relevant slot then yields a shared query and result oracle together with three matched prefixes (correct, explicit, and implicit). The workflow and clean trace define the fixed evaluation environment $E_i$, in which the evaluated LLM continues from each prefix to produce a complete rollout.}
    \label{fig:benchmark-infrastructure}
\vspace{-8pt}
\end{figure}

The infrastructure has a shared start followed by two sibling construction branches:
\[
    (W_i,x_i)\longmapsto\tau_i,
    \qquad
    \tau_i\longmapsto D_i^k=(q_i,p_i^k,c_i),
    \qquad
    (W_i,\tau_i)\longmapsto E_i.
\]
The data branch constructs the paired prefixes initializing an agent, whereas the environment branch independently constructs the world where the agent continues acting (Fig.~\ref{fig:benchmark-infrastructure}).

\paragraph{Workflow trace reconstruction.}
To avoid polluting the internal platform data, which may record the workflow execution result, we first post-process every selected workflow into a deployable replica by mocking out reporting and other side-effecting nodes and deploy the modified workflow back onto the platform. 
Then, a local central controller connects the log and workflow to provide an execution trace. It initially maps the production log record to the start schema of the deployed replica, producing the input $x_i$. It then invokes the deployed workflow through the platform API with $x_i$ and retrieves the full execution trace of that run, which records every executed node with its arguments, outputs, status, and data dependencies.
We retain only executions that complete successfully and contain full information needed to reconstruct the workflow state. 
The retained $\tau_i$ is therefore the clean execution record shared by the two downstream branches.

\paragraph{Paired prefix construction.}

\textbf{Slot selection.}
We align the workflow graph with $\tau_i$ and obtain an execution graph whose edges carry the values propagated between nodes. We then search this graph along
\(
\text{producer}\rightarrow s_i\rightarrow\text{consumer}\rightarrow\text{downstream result}
\): an intermediate slot $s_i$ qualifies when it carries business-relevant content, is consumed downstream, and plausibly affects the task outcome. For each retained slot, we record its producer and consumer and determine the locus at which the intervention is realized, either in the producer observation (\texttt{tool\_output}) or in a downstream tool argument (\texttt{tool\_input}). A single execution yields multiple such slots, and each retained slot defines one task through the prefix built around it.

\textbf{Task construction.}
This stage converts each retained slot into a task including the task query and corresponding three paired prefixes.
\begin{enumerate}[label=(\roman*),leftmargin=*,itemsep=0pt,topsep=1pt,parsep=0pt]
    \item \textbf{Three conditions at the same slot.} We truncate the common trace at the same slot and render the three prefixes $p_i^{\mathrm{correct}}$, $p_i^{\mathrm{explicit}}$, and $p_i^{\mathrm{implicit}}$ (Sec.~\ref{sec:formulation}), with neutral historical rationales and no corruption marker. 
    The correct condition preserves the recorded value. 
    The explicit condition replaces access to that value with an overt failure, such as an empty argument or an error-bearing observation. 
    The implicit condition obtains its value in two steps. First, it selects an applicable error type from the taxonomy and constructs a substitute that is type- and schema-compatible, remains locally plausible, and changes the task-relevant meaning of the slot. Depending on the error type and carrier, candidates come from compatible values observed in other cases or from a type-specific controlled transformation. Second, the selected substitute is written into the slot at its predetermined locus: the producer observation under \texttt{tool\_output}, or a downstream tool argument under \texttt{tool\_input}. 
    \item \textbf{Query and oracle.} From the clean trace, we extract the correct value of $s_i$ and construct the result constraint $c_i$ that a correct final answer must satisfy.
    We then generate or refine $q_i$ so that it is designed to depend on this value: answers grounded in the gold and polluted values should be distinguishable under $c_i$.
\end{enumerate}

\textbf{Verification.}
Before delivery, each candidate must preserve prefix executability and structural compatibility. Four additional gates then check the intervention itself. \textbf{(1) Plausibility.} The implicit value must remain schema-valid and locally reasonable while differing from the recorded value in task-relevant meaning. \textbf{(2) Explicit-failure clarity.} The explicit condition must expose an unambiguous failure signal. \textbf{(3) Non-leakage.} Neither the query nor the prefix may reveal that an intervention was applied. \textbf{(4) Discriminability.} The query remains solvable, and answers grounded in the gold value and in the polluted value must remain distinguishable under $c_i$. The surviving artifacts are delivered in the formulation's schema $D_i^k=(q_i,p_i^k,c_i)$, with all three conditions sharing the same query and result constraint.
Providing every model with the same recorded execution before the affected value deliberately reduces variation in reaching that point, so the matched conditions focus on how agents handle the evidence they receive there.

\paragraph{Evaluation environment construction.}
\label{sec:data:env}
In parallel with prefix construction, we extract the callable tool set $\mathcal{T}_i$ from $W_i$ and build one replay world $R_i$ for that workflow by integrating the recorded responses of all collected traces. 
\begin{equation}
    (W_i,\tau_i)\longmapsto E_i(a_t)=
    \begin{cases}
        R_i(a_t), & a_t\in\operatorname{dom}(R_i),\\
        \operatorname{Live}_{\mathcal{T}_i}(a_t), & \text{otherwise}.
    \end{cases}
    \label{eq:data-env}
\end{equation}
Under this interface, a covered call returns the recorded response from $R_i$, while an unseen valid call is dispatched to the live implementation in $\mathcal{T}_i$ deployed on the internal industrial workflow platform.
\benchname{} changes only the condition-specific prefix, not $E_i$; because the replay world is built from the clean trace, re-querying a recorded producer returns the corresponding unpolluted response, giving the agent a direct path to recover reliable evidence and produce a correct final answer.
Every instance is therefore recoverable by design, isolating whether agents use an available verification path; faults that persist across re-queries are discussed in Appendix~\ref{app:limitations}.

\subsection{Dataset Statistics}
\label{sec:data:stats}

The final \benchname{} release contains 167 paired units, each defined by a unique slot--locus pair and expanded into correct, explicit, and implicit conditions. This yields 501 instances, with 167 instances per condition and complete triplets throughout. Nine workflows contribute data, and the release covers six of the seven taxonomy error types because temporal staleness has no retained instance. Of the 167 units, 147 intervene at \texttt{tool\_output}; the remaining 20 intervene at \texttt{tool\_input}. Detailed distributions are reported in Appendix~\ref{app:dataset-distribution}.


\section{Experiments}
\label{sec:exp}


We evaluate 10 recent LLMs in three groups: two closed frontier models (Claude-Opus-4.8 and GPT-5.5), five open-weight frontier models (GLM-5.2, Kimi-K3, DeepSeek-V4-Pro, Hy3, and Qwen3.5-397B), and three smaller open-weight models (Qwen3.6-35B, Qwen3.5-35B, and Qwen3.5-122B). Each model runs all 501 instances at temperature 0 with a 25-step budget. Rollouts invalidated by environment-side tool anomalies are excluded. The evaluation environment follows Section~\ref{sec:data:env}, and the rollout and judging protocol is detailed in Appendix~\ref{app:evaluation-protocol}, and a human validation of the Result and Noticed labels is reported in Appendix~\ref{app:human-validation}. Following Section~\ref{sec:formulation}, we report condition-specific RSR and Noticed, Retention, and the paired gap: the paired explicit-minus-implicit RSR difference over slots observed in both conditions.

\subsection{Main Results}
\label{sec:exp:main}

Table~\ref{tab:main} shows that implicit errors substantially degrade task performance across all evaluated models and, in most cases, remain unnoticed. 
Appendix~\ref{app:worked-refetch-example} provides two end-to-end cases.

\begin{table*}[!htbp]
\centering
\small
\setlength{\tabcolsep}{2pt}
\caption{Results across the three matched conditions, grouped by model class and ordered within each group by implicit-condition Noticed. The Noticed gap is explicit minus implicit. \textbf{Bold} marks the best value in each column. 
}
\resizebox{0.85\textwidth}{!}{%
\begin{tabular}{llcccccccc}
\toprule
& & \multicolumn{3}{c}{Noticed} & \multicolumn{3}{c}{RSR} & \multicolumn{2}{c}{Resilience} \\
\cmidrule(lr){3-5}\cmidrule(lr){6-8}\cmidrule(lr){9-10}
Category & Model & Implicit $\uparrow$ & Explicit $\uparrow$ & Gap
& Correct $\uparrow$ & Implicit $\uparrow$ & Explicit $\uparrow$
& Retention $\uparrow$ & Paired gap \\
\midrule
Closed frontier & Claude-Opus-4.8 & \textbf{0.336} & 0.579 & 0.243 & 0.765 & \textbf{0.403} & 0.507 & \textbf{0.526} & \textbf{+0.089} \\
& GPT-5.5 & 0.280 & \textbf{0.594} & 0.314 & 0.716 & 0.363 & 0.526 & 0.507 & +0.185 \\
\midrule
Open-weight frontier & Kimi-K3 & 0.263 & 0.590 & 0.327 & 0.784 & 0.302 & 0.555 & 0.385 & +0.228 \\
& GLM-5.2 & 0.217 & 0.487 & 0.270 & 0.805 & 0.277 & 0.507 & 0.345 & +0.243 \\
& Hy3 & 0.215 & 0.494 & 0.278 & 0.806 & 0.310 & 0.552 & 0.385 & +0.236 \\
& DeepSeek-V4-Pro & 0.212 & 0.468 & 0.255 & \textbf{0.827} & 0.327 & 0.533 & 0.395 & +0.217 \\
& Qwen3.5-397B & 0.210 & 0.477 & 0.267 & 0.809 & 0.320 & 0.543 & 0.396 & +0.220 \\
\midrule
Smaller open-weight & Qwen3.6-35B & 0.248 & 0.512 & 0.264 & 0.823 & 0.335 & 0.531 & 0.408 & +0.210 \\
& Qwen3.5-35B & 0.211 & 0.512 & 0.301 & 0.791 & 0.340 & 0.534 & 0.429 & +0.208 \\
& Qwen3.5-122B & 0.208 & 0.506 & 0.299 & 0.784 & 0.302 & \textbf{0.564} & 0.385 & +0.265 \\
\bottomrule
\end{tabular}
}
\vspace{-8pt}
\label{tab:main}
\end{table*}

\paragraph{F1: Silent errors substantially degrade task performance across models.}\label{sec:exp:f1}
Across all ten models, implicit errors reduce RSR from 0.716--0.827 in the correct condition to 0.277--0.403, yielding Retention of only 0.345--0.526 (Table~\ref{tab:main}). 
This loss is tied to the absence of an observable failure signal: when the corresponding failure is explicit, RSR rises to 0.507--0.564, producing a positive explicit-minus-implicit paired gap for every model (+0.089 to +0.265; +0.214 pooled). 
Stronger task performance does not eliminate this vulnerability. Correct-condition RSR spans only 0.111 across models, whereas Retention spans 0.182; closed frontier models have higher mean Retention than open-weight models (0.517 versus 0.391), yet even the best model preserves barely half of its correct-condition performance. Thus, susceptibility to silent errors is widespread rather than confined to models with weak baseline task ability.

\paragraph{F2: Models rarely notice silent errors or refetch their producers.}\label{sec:exp:f2}
\begin{wraptable}[7]{R}{0.31\linewidth}
\vspace{-15pt}
\centering
\caption{Latest valid producer observation by Noticed in the implicit condition.}
\label{tab:f2f3}
\vspace{-5pt}
\scriptsize
\setlength{\tabcolsep}{1.2pt}
\renewcommand{\arraystretch}{0.95}
\begin{tabular}{@{}lrrrr@{}}
\toprule
Noticed & $n$ & Gold & Polluted & Other \\
\midrule
No & 1198 & 0.3\% & 99.6\% & 0.1\% \\
Yes & 377 & 23.3\% & 57.8\% & 18.8\% \\
\bottomrule
\end{tabular}
\vspace{-10pt}
\end{wraptable}
Across models, implicit-condition Noticed ranges only from 0.208 to 0.336 (Table~\ref{tab:main}), so most silent errors go unrecognized. Table~\ref{tab:f2f3} shows that recognition helps but is not sufficient: the latest producer observation matches the gold value in 0.3\% of Unnoticed rollouts and 23.3\% of Noticed rollouts, while 57.8\% of Noticed rollouts still match the polluted value.
To examine whether recognition leads to active verification, we track \emph{producer refetch}, defined as re-invoking the tool that produced the affected value. Among scored Noticed rollouts, RSR is 0.387 without a producer refetch and 0.655 with one.
GLM-5.2 illustrates this gap: its Noticed rate is 0.217, but its producer-refetch rate is only 0.071 and its Retention is the lowest at 0.345. Across the eight open-weight models, producer-refetch rate is positively but not significantly associated with Retention ($r=0.491$, $p=0.217$). These results identify two bottlenecks: models rarely notice silent errors, and noticing one often does not lead to refetching its source.

\paragraph{F3: Producer refetch does not ensure recovery.}\label{sec:exp:f3}


To test whether producer refetch restores the correct evidence, the state judge compares the latest valid producer observation with the gold and polluted values. Among the 147 implicit-condition rollouts with a producer refetch, 87 (59.2\%) match the gold value and 60 (40.8\%) match neither value; none match the polluted value. Of the 87 matches, 85 are obtained in one re-query and the remaining two require multiple rounds.
Among Noticed rollouts with a producer refetch, about one third still fail the task (RSR 0.655; F2).

To understand why refetching does not ensure recovery, we manually inspect the 60 trajectories that did not match value, focusing on unsolved rollouts in which the re-query nonetheless returns usable evidence. After usable evidence is acquired, we identify three recurring patterns. First, \emph{reasoning overwrite} occurs when a model initially uses recovered evidence but later replaces it with a self-imposed heuristic. In one trajectory, an agent reconstructs the correct timestamps and uses them in the next RPC call, but then redefines the ``opening warm-up'' boundary and selects the wrong final comment. 
Second, \emph{lossy selection} occurs when a model alters recovered candidates while choosing among them. One agent merges two recovered topic clusters despite an instruction to select one, while another replaces the retrieved topic with a plausible but unsupported reformulation. Third, \emph{unsupported synthesis} occurs when a model uses corrected evidence but adds claims absent from it. Agents incorporate corrected profile or interest labels but add unsupported viewing-duration, interaction, or presentation claims.
Across these patterns, usable evidence is obtained, but the correction does not carry through to the final answer.


\section{Discussion}
\label{sec:discussion}

The main results separate silent-error handling into recognition, use of an available verification path, and integration of the resulting evidence into final recovery. This section asks three follow-up questions: whether the resulting performance loss corresponds to changed e-commerce task outputs (D1), where the vulnerability concentrates (D2), and how far a light check can go (D3).

\paragraph{D1: Do silent errors change e-commerce task outputs?}\label{sec:discussion:d1}
Not every corrupted task-relevant value changes the downstream e-commerce output.
We therefore use an LLM judge to identify slots where answering with the gold versus the polluted value would change the business conclusion (Appendix~\ref{app:evaluation-protocol}) and recompute Retention on this subset. Table~\ref{tab:d1:overall} compares it with the full benchmark, while Table~\ref{tab:d1:error-type} reports the output-change rate by error type.

\begin{wraptable}[15]{r}{0.38\linewidth}
\vspace{-10pt}
\centering
\caption{E-commerce task-output impact and result robustness.}
\label{tab:d1}
\vspace{2pt}
\scriptsize
\setlength{\tabcolsep}{1.6pt}
\renewcommand{\arraystretch}{0.94}
\captionsetup[subtable]{font=scriptsize,skip=2pt}
\begin{subtable}{\linewidth}
\centering
\begin{tabular*}{\linewidth}{@{\extracolsep{\fill}}lrrr@{}}
\toprule
Subset & Slots & Share & Ret. \\
\midrule
All & 167 & 100.0\% & 0.416 \\
Output-changing & 146 & 87.4\% & 0.408 \\
\bottomrule
\end{tabular*}
\caption{Overall output impact.}
\label{tab:d1:overall}
\end{subtable}

\vspace{3pt}
\begin{subtable}{\linewidth}
\centering
\begin{tabular*}{\linewidth}{@{\extracolsep{\fill}}lrr@{}}
\toprule
Error type & Slots & Change rate \\
\midrule
Cat. granularity & 17 & 100.0\% \\
Cat. sibling & 20 & 95.0\% \\
Profile attribute & 35 & 94.3\% \\
Semantic synonym & 40 & 92.5\% \\
Entity near-duplicate & 23 & 82.6\% \\
Numeric bucket & 32 & 65.6\% \\
\bottomrule
\end{tabular*}
\caption{Output-change rate by error type.}
\label{tab:d1:error-type}
\end{subtable}
\vspace{-10pt}
\end{wraptable}

The gold and polluted values are judged to lead to different task outputs for 146 of 167 slots (87.4\%). 
Depending on the workflow, this means a different product recommendation, live-assistant intervention, topic suggestion, or other task result. The change rate is at least 82.6\% for every non-numeric error type and 65.6\% for numeric bucket shifts. 
In a typical non-changing numeric case, average watch times of 64 and 77 seconds both clear a 60-second threshold and therefore produce the same downstream result. Restricting evaluation to the 146 output-changing slots changes mean Retention only from 0.416 to 0.408. The near-identical Retention shows that F1's performance loss is not driven by factual differences that leave downstream behavior unchanged; it persists, and is slightly larger, when silent errors alter the e-commerce action or result.

\paragraph{D2: Damage concentrates by error type and slot position.}\label{sec:discussion:d2}
Damage varies across both silent-error types and slot positions: it is greatest for category errors (Table~\ref{tab:d2:error-type}) and for slots that occur early in the workflow trace (Table~\ref{tab:d2:position}).


\textbf{Error type.} Category errors are the most damaging error types in \benchname{}: category granularity mismatch is hardest, category sibling drift is next, and numeric bucket shift is most noticeable but remains destructive (Table~\ref{tab:d2:error-type}. One possible explanation is that in the benchmark's recoverable environment, a wrong number or ID is checked against the producer response, whereas a wrong category appears anomalous only in context. For example, identifying ``blockchain'' as an implausible interest requires knowing the user watches local sports; the category value alone carries no anomaly.

\textbf{Slot position.} Early slots have the highest Noticed rate (0.264) but the lowest implicit-condition RSR (0.309) and Retention (0.399; Table~\ref{tab:d2:position}). One plausible explanation is that an early checkpoint yields a shorter prefix with fewer inherited observations, making its corruption easier to notice. At the same time, the agent must complete a longer downstream path, giving the polluted value---or an incomplete recovery---more opportunities to affect later steps. 

\begin{wraptable}[15]{l}{0.38\linewidth}
\vspace{-10pt}
\centering
\caption{Damage localization. Slots are paired units; Not. and RSR are measured in the implicit condition.}
\label{tab:d2}
\vspace{2pt}
\scriptsize
\setlength{\tabcolsep}{1.6pt}
\renewcommand{\arraystretch}{0.94}
\captionsetup[subtable]{font=scriptsize,skip=2pt}
\begin{subtable}{\linewidth}
\centering
\begin{tabular*}{\linewidth}{@{\extracolsep{\fill}}lrccc@{}}
\toprule
Error type & Slots & Not. & RSR & Ret. \\
\midrule
Category granularity & 17 & 0.099 & 0.140 & 0.180 \\
Category sibling & 20 & 0.121 & 0.199 & 0.292 \\
Numeric bucket & 32 & 0.342 & 0.224 & 0.297 \\
Semantic synonym & 40 & 0.231 & 0.307 & 0.382 \\
Profile attribute & 35 & 0.268 & 0.479 & 0.562 \\
Entity near-duplicate & 23 & 0.258 & 0.507 & 0.608 \\
\bottomrule
\end{tabular*}
\caption{Damage by silent-error type.}
\label{tab:d2:error-type}
\end{subtable}

\vspace{3pt}
\begin{subtable}{\linewidth}
\centering
\begin{tabular*}{\linewidth}{@{\extracolsep{\fill}}lrccc@{}}
\toprule
Slot position & Slots & Not. & RSR & Ret. \\
\midrule
Early & 128 & 0.264 & 0.309 & 0.399 \\
Mid & 26 & 0.175 & 0.384 & 0.452 \\
Late & 13 & 0.122 & 0.395 & 0.470 \\
\bottomrule
\end{tabular*}
\caption{Damage by slot position.}
\label{tab:d2:position}
\end{subtable}
\vspace{-10pt}
\end{wraptable}
The positional trend therefore separates detection from final success: from early to late, Noticed falls, while implicit-condition RSR and Retention rise. This pattern makes upstream verification especially important.

\begin{figure}[H]
\centering
\includegraphics[width=\linewidth]{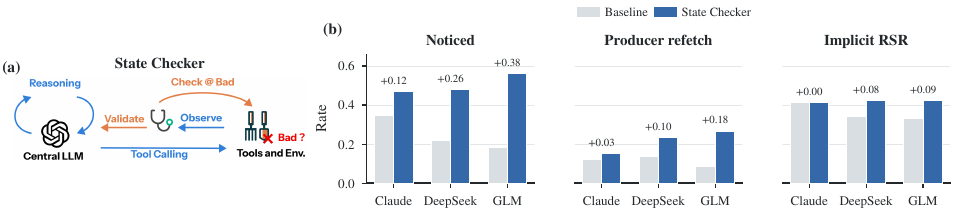}
\caption{\textbf{(a)} State Checker adds a reliability check before a tool result is used. \textbf{(b)} Noticed, producer-refetch, and implicit-condition RSR with/without State Checker on paired implicit slots.}
\label{fig:d3}
\vspace{-10pt}
\end{figure}

\paragraph{D3: A lightweight intervention for handling silent errors.}\label{sec:discussion:d3}
F2 shows that blind trust in inherited tool results is the dominant failure pattern, suggesting an additional reliability check during execution.
We test State Checker, which asks the model to treat every tool result as potentially unreliable and, before using it, conduct an additional step to reflect on whether the result is trustworthy and should be verified.
The underlying tools and benchmark instances remain unchanged. Claude-Opus-4.8, DeepSeek-V4-Pro, and GLM-5.2 each rerun half of their slots (Appendix~\ref{app:evaluation-protocol}) with the additional check, paired against the same slots without it.
State Checker increases Noticed and producer-refetch rates, but produces only modest gains in implicit-condition RSR (Figure~\ref{fig:d3}).

\noindent State Checker changes behavior in all three models, with larger increases in Noticed and producer refetch than in final RSR. The clearest behavioral shift is in GLM-5.2, whose producer-refetch rate more than triples; by contrast, Claude-Opus-4.8 gains no RSR despite noticing more errors.


State Checker makes evidence-seeking behavior steerable: an explicit reliability check increases recognition and use of the available verification path. The smaller RSR gains locate the remaining bottleneck in evidence integration and final task success, while the surfaced contradictions improve observability for a human or downstream controller.
State Checker relies on the unpolluted verification path in \benchname{}'s single-site corruption setting. Persistent or system-wide tool faults may invalidate this path, requiring independent sources, tool substitution, redundancy, or human check.

\section{Conclusion}
\label{sec:conclusion}
We study silent errors in agentic tasks, where a plausible but incorrect tool result enters the execution history without an explicit failure signal and propagates through subsequent steps.
We formulate a task that measures whether an agent can tolerate this corrupted state and still complete the original task, and introduce \benchname{}, a benchmark built from 167 corruption slots across nine industrial e-commerce workflows grounded in real service logs.
Across 10 recent LLMs, silent errors roughly halve task success and usually remain unnoticed; recovery remains incomplete even when models question the affected evidence and re-query its producer.
Further analysis shows that these errors often alter downstream e-commerce outputs, with category errors and early-stage corruptions causing the greatest damage.
Beyond the benchmark, we provide a simple reliability check that increases error detection and producer refetching more than final task success.

\clearpage
\section*{Author Statements}

\subsection*{AI Use Statement}

In this work, we used generative AI tools in four roles. First, the authors wrote the initial draft of every manuscript section, after which AI tools were used to polish the language and improve clarity. Second, AI assisted benchmark construction by helping screen candidate workflows and production-log records (Appendix~\ref{app:data-sources}) and by serving as model components during workflow execution. The authors manually reviewed every workflow selected through this process. Third, GLM-5 serves as the automatic judge of task success and observation state, and labels whether a corruption would change the business conclusion (Appendix~\ref{app:evaluation-protocol}). Fourth, AI assisted with coding. The authors reviewed and validated all AI-assisted code, screening decisions, workflow outputs, and generated benchmark artifacts before use.

\subsection*{Ethics Statement}

The workflows used in this study were deployed in production, and the logs were generated by their online operation. Before any log-derived data were provided to language models, sensitive identifiers such as user IDs were anonymized; benchmark runs used only anonymized data.

Benchmark data and generated outputs were isolated from production storage: no benchmark inputs, intermediate states, or outputs were reported to or written back to company production databases.

\subsection*{Reproducibility Statement}

The paper documents the task, data sources, benchmark construction, evaluation protocol, dataset distribution, and auditable workflow examples needed to understand and assess the reported results.

The code is subject to an internal company review that may take several months. We will release it if approval is granted; otherwise, it will remain unavailable.

\bibliography{iclr2027_conference}
\bibliographystyle{iclr2027_conference}

\appendix
\section{Workflow and Log Data Sources}
\label{app:data-sources}

This appendix documents two data-source stages: selecting the workflow collection $\mathcal{W}$ and mapping production logs to workflow inputs that yield eligible traces. It also records the resulting golden workflows. Workflow identifiers are anonymized and stable across the released artifacts.

\paragraph{Workflow selection: structural validity.}
We inspected e-commerce applications at the DSL level and retained complete deterministic DAGs with explicit control flow. We excluded graphs containing autonomous agentic nodes, atomic retrieval wrappers, short utility graphs, AB-test fragments, dangling references, or incomplete subgraphs. Fixed tool calls, read-only requests, code and LLM transformations, and explicit branch or loop nodes remained eligible.

\paragraph{Workflow selection: task diversity.}
Because the application library contains multiple versions of many task families, we deduplicated each family, selected a representative complete version, and formed a set spanning different business stages, task objectives, and data dependencies. These two workflow checks yielded nine golden workflows.

\paragraph{Log selection and trace retention.}
For each selected workflow, we mapped each candidate production-log record to the required start inputs and executed it on a structure-preserving replica. We selected a log only when its execution yielded a complete node-level trace with sufficient process depth and at least one business-semantic intermediate state that was produced upstream, consumed downstream, and locatable at a producer or consumer checkpoint. The corresponding trace was retained with the selected log. For safe trace collection, the replicas retained node identities, graph edges, data references, read-only retrieval, deterministic computation, and model-inference nodes, while writes, reporting calls, locks, memory updates, and external sends were neutralized or guarded.

Table~\ref{tab:workflow-inventory} summarizes the selected workflows, their log-derived execution contexts, graph sizes, and resulting final units. Graph size reports the total nodes and statically counted tool- or model-facing interfaces in each collection replica. The interface count is descriptive and precedes trace-level eligibility checks. A \emph{final unit} is one retained slot--locus pair before expansion into the three conditions.

\begingroup
\scriptsize
\setlength{\tabcolsep}{2pt}
\begin{longtable}{L{0.06\linewidth}L{0.25\linewidth}L{0.36\linewidth}L{0.14\linewidth}L{0.08\linewidth}}
\caption{The nine golden workflows used by \benchname{}. Log-derived inputs are described by semantic role rather than internal field names. Graph size is reported as total nodes / static tool-model interfaces.}
\label{tab:workflow-inventory}\\
\toprule
ID & Task abstraction & Log-derived execution context & Graph size & Final units \\
\midrule
\endfirsthead
\multicolumn{5}{l}{\small\itshape Table~\ref{tab:workflow-inventory} continued.}\\
\toprule
ID & Task abstraction & Log-derived execution context & Graph size & Final units \\
\midrule
\endhead
\midrule
\multicolumn{5}{r}{\small\itshape Continued on next page.}\\
\endfoot
\bottomrule
\endlastfoot
wf01 & Product recommendation from user needs, creator profiles, and selection history & Anonymized viewer, creator, and request context. A task query is constructed from the available profile context & 55 / 14 & 11 \\
wf02 & Live-assistant scheduling and intervention selection & Anonymized session, creator, task, and configuration context & 58 / 11 & 22 \\
wf04 & Pre-session script generation from creator and audience context & Anonymized creator identity. Profiles and historical content are retrieved by read-only nodes & 22 / 11 & 15 \\
wf05 & Real-time product-presentation coaching & Anonymized session, creator, task, and available configuration context & 47 / 12 & 17 \\
wf06 & Cross-domain intent and interest modeling & An anonymized user identity passing data-availability screening. Behavioral and profile records are retrieved downstream & 52 / 14 & 52 \\
wf07 & Post-session summary over content, interaction, and commerce signals & Anonymized session, creator, request, and a session-derived time window & 56 / 10 & 15 \\
wf08 & Topic suggestion from historical videos and comments & Anonymized creator, video or session, and request context & 21 / 5 & 12 \\
wf09 & Extraction of reusable presentation patterns from strong sessions & Anonymized session, creator, request, and analysis window & 37 / 5 & 9 \\
wf10 & Periodic perception of the current live-session state & Anonymized session, creator, task, and available configuration context & 46 / 9 & 14 \\
\midrule
\multicolumn{3}{l}{\textbf{Total: nine golden workflows}} & \textbf{394 / 91} & \textbf{167} \\
\end{longtable}
\endgroup

\section{Silent-Error Taxonomy and Literature Evidence}
\label{app:taxonomy-evidence}

We synthesize 15 papers spanning agent and tool failures, data quality, and factuality. Two authors independently coded the error phenomena and boundaries described by each source, mapped overlapping concepts to a common vocabulary, and reconciled the mappings. The resulting taxonomy describes the semantic form of a plausible but incorrect intermediate value. Injection location is recorded separately.

\subsection{Operational Definitions}

The main taxonomy contains two semantic groups and seven leaves (Table~\ref{tab:taxonomy-leaves}). The first group changes which object or category is referenced; the second preserves the object while changing one of its attributes or values. Each leaf requires a schema-valid value whose task-relevant meaning differs from the recorded value.

\begin{table}[H]
\centering
\caption{Operational definitions and boundaries of the seven silent-error leaves.}
\label{tab:taxonomy-leaves}
\vspace{3pt}
\scriptsize
\setlength{\tabcolsep}{3pt}
\begin{tabular}{L{0.08\linewidth}L{0.25\linewidth}L{0.31\linewidth}L{0.27\linewidth}}
\toprule
ID & Leaf & Operational definition & Primary boundary \\
\midrule
L1.1 & \texttt{category\_\allowbreak sibling\_\allowbreak drift} & Replace a category with a nearby category at the same hierarchy level. & Parent--child substitutions belong to L1.2. \\
L1.2 & \texttt{category\_\allowbreak granularity\_\allowbreak mismatch} & Replace a category with an ancestor or descendant, making its scope too broad or too narrow. & Same-level substitutions belong to L1.1. \\
L1.3 & \texttt{entity\_\allowbreak near\_\allowbreak duplicate} & Replace the intended entity with another real, similar entity from the candidate domain. & If entity identity is preserved and only an attribute changes, use L2.1. \\
L2.1 & \texttt{profile\_\allowbreak attribute\_\allowbreak drift} & Preserve entity identity but replace a profile, preference, label, or other attribute with another valid value. & Whole-entity replacement belongs to L1.3. \\
L2.2 & \texttt{numeric\_\allowbreak bucket\_\allowbreak shift} & Move a numeric value or interval to a neighboring valid range. & Malformed, impossible, or plainly out-of-range values are explicit failures. \\
L2.3 & \texttt{semantic\_\allowbreak synonym\_\allowbreak drift} & Substitute text that appears semantically close but changes the business interpretation used downstream. & Meaning-preserving paraphrases are not errors. \\
L2.4 & \texttt{temporal\_\allowbreak staleness} & Reuse a value that was valid previously but is no longer correct at the task time. & A contemporaneous neighboring numeric value belongs to L2.2. \\
\bottomrule
\end{tabular}
\end{table}

The injection locus is annotated independently of the semantic leaf and takes one of two values: \texttt{tool\_output}, where the substitute is written into the producer observation, or \texttt{tool\_input}, where it is written into a downstream tool argument.

\subsection{Literature Grounding}

\paragraph{Silentness and exclusion boundaries.}
\emph{Tools Fail} \citep{sun-etal-2024-tools} establishes the central operational criterion: a tool output may be schema-valid and wrong without raising an exception. The \emph{Tool-Augmented LLM Failure Taxonomy} \citep{11081716} reinforces the distinction between explicit and silent failures. The \emph{LLM Hallucination Survey} \citep{Huang_2025} and \emph{Where LLM Agents Fail} \citep{zhu2025llmagentsfaillearn} further separate entity, context, profile, memory, and parameter errors, motivating value-level interventions while excluding failures that arise only from the agent's internal reasoning or memory. Other taxonomies delimit the scope from adjacent phenomena: the \emph{Multi-Agent Failure Taxonomy} \citep{cemri2025multiagentllmsystemsfail} covers behavioral and system failures, $\tau$-bench \citep{yao2025taubench} emphasizes domain-policy compliance, CRMArena-Pro \citep{huang2026crmarenapro} adds confidentiality failures, and \emph{Reasoning Error Localization} \citep{tyen2024llmsreasoningerrorscorrect} distinguishes recognizing an error from correcting its downstream effects. Together, these sources define silent error as an incorrect intermediate value rather than a planning, collaboration, policy, confidentiality, or reasoning failure.

\paragraph{Semantic distinctions.}
Data-quality and factuality research supplies the distinctions used to organize the seven leaves. The \emph{Dirty Data Taxonomy} \citep{3e4f9299fc754457883b482e07e49636} separates category, value, relation, and stale-data errors, supporting category-level substitutions, attribute and numeric changes, and temporal staleness. \emph{Faithfulness versus Factuality} \citep{maynez-etal-2020-faithfulness} frames an error as a contradiction with its source evidence, providing a common criterion across all leaves. The taxonomy of \emph{Factual Errors in Summarization} \citep{tang-etal-2023-understanding} distinguishes entity, attribute, numeric, and temporal errors, which map naturally to entity near-duplicates, profile-attribute drift, numeric-bucket shifts, and temporal staleness.

\paragraph{Specialized leaves and loci.}
Several sources refine particular boundaries. \emph{Butterfly Effects in Toolchains} \citep{xiong-etal-2025-butterfly} shows how incorrect tool arguments propagate downstream, motivating \texttt{tool\_input} as an intervention locus in addition to \texttt{tool\_output}. Work on \emph{Duplicate Record Detection} \citep{1191547.1191739} motivates the distinction between real near-duplicate entities and superficially similar text. \emph{FreshLLMs} \citep{vu-etal-2024-freshllms} grounds temporal staleness in answers that change over time. Finally, \emph{LM-Emulated Sandbox Risks} \citep{ruan2024identifyingriskslmagents} helps exclude system and safety failures from the semantic taxonomy.

Taken together, the agent and tool literature defines when an error is silent and what lies outside scope, while the data-quality and factuality literature supplies the semantic forms of the corrupted value. We adapt these distinctions into schema-valid, single-slot workflow interventions; the semantic leaf describes what changed, and the locus records where that change enters the trajectory.

\section{Evaluation Protocol Details}
\label{app:evaluation-protocol}

This appendix specifies the rollout, judging prompts, and process measures used in evaluation.
All prompts, trajectories, and case-study contents in this paper are English translations.

\paragraph{Evaluation flow.}
Each rollout follows four stages:
\begin{enumerate}[label=(\roman*),leftmargin=*,itemsep=1pt,topsep=2pt,parsep=0pt]
    \item \textbf{Initialize.} For benchmark unit $i$ and condition $k\in\{\mathrm{correct},\mathrm{implicit},\mathrm{explicit}\}$, the query is $q_i$, the continuation environment is $E_i$, and the condition-specific prefix $p_i^k$ initializes the agent history as $h_{0,i}^k=p_i^k$. The correct prefix preserves the recorded value, the implicit prefix substitutes a silently incorrect value, and the explicit prefix presents an overt failure. The agent receives the task instruction and task inputs, followed by the prefix rendered as a numbered record of thoughts, actions, and observations.
    \item \textbf{Execute.} The agent follows Eq.~\ref{eq:agent-loop} for at most 25 steps. At each step, it outputs a thought and a JSON action that either calls a tool with arguments or stops with a final answer. Each condition uses a newly initialized copy of the same workflow-specific tool registry and case-specific replay snapshot. Tool names are normalized through workflow aliases. For each call, $E_i$ first searches for an argument-matched recorded response and returns it if found; repeated unmatched calls may consume the next unconsumed response in recorded order, while calls absent from or exhausted in the snapshot fall back to the live read-only toolbox. Replay hits, misses, live fallbacks, and observation lengths are recorded for audit.
    \item \textbf{Judge.} GLM-5 at temperature 0 independently applies the state-judge and result-judge prompts below. The same model also judges whether an agent that faithfully uses the gold versus the polluted value would reach a different business conclusion, which defines the output-changing slots in D1 (Section~\ref{sec:discussion}). 
    \item \textbf{Measure.} Producer refetch and Noticed are computed from the trajectory independently of both judge outputs.
\end{enumerate}

\paragraph{Rollout counts and exclusions.}
Across the ten models, each condition has 1,670 scheduled rollouts; in total, 73 scheduled rollouts produced no valid record. A rollout is excluded if any of its observations contains a tool-execution error message, which indicates an environment-side anomaly; this rule removes 201 rollouts. The retained rollouts number 1,599, 1,575, and 1,562 for the correct, implicit, and explicit conditions. Noticed is computed on all retained rollouts, and RSR on the 1,576, 1,555, and 1,543 retained rollouts that received a result judgment.

\paragraph{Judge prompts.}
Braced fields denote values filled separately for each benchmark instance.

\textbf{State judge.}
The state judge determines whether the latest valid producer observation recovers the affected evidence.
\begin{quote}
\footnotesize\raggedright
You are a strict evaluator. An intermediate slot may have been corrupted.\\
\textbf{Slot:} \texttt{\{slot\}}; \textbf{condition:} \texttt{\{condition\}}.\\
\textbf{Gold:} \texttt{\{gold\_value\}}.\\
\textbf{Polluted:} \texttt{\{polluted\_value\}}.\\
\textbf{Latest observation after the agent re-invoked the producer:} \texttt{\{latest\_observation\}}.\\
Classify the value as \texttt{gold}, \texttt{polluted}, \texttt{neither}, or \texttt{unknown}. Differences in formatting do not change semantic equivalence. Return only JSON with fields \texttt{used} and \texttt{reason}.
\end{quote}
The outputs \texttt{gold}, \texttt{polluted}, and \texttt{neither} correspond to Recovered, Residual, and Neither observation states, respectively; \texttt{unknown} indicates insufficient evidence for attribution.

\textbf{Result judge.}
The result judge determines whether the final answer completes the original user task.
\begin{quote}
\footnotesize\raggedright
You are a strict evaluator. Decide only whether the agent's final answer solves the user task.\\
\textbf{User task:} \texttt{\{user\_task\}}.\\
\textbf{Correct reference evidence:} \texttt{\{correct\_reference\_evidence\}}.\\
\textbf{New observations:} \texttt{\{new\_observations\}}.\\
\textbf{Final answer:} \texttt{\{final\_answer\}}.\\
Check whether every task-critical claim is supported by either the reference evidence or a new observation. A contradiction unsupported by new evidence, an incomplete answer, or an error termination is unsolved. Return only JSON with fields \texttt{solved} and \texttt{reason}.
\end{quote}
The \texttt{solved} output defines RSR. The state and result judgments are independent: recovering the affected evidence does not by itself guarantee a correct final answer.

\paragraph{Process measures.}
\emph{Producer refetch} records whether the agent invokes the producer of the affected value again after that value appears in the prefix. \emph{Noticed} is true when the reasoning explicitly signals a related content or result anomaly and is followed by a remedial action, such as refetching the producer, consulting a previously unused corroborating tool, or issuing an explicit refetch. 

\begin{table}[H]
\centering
\caption{Objects evaluated by the process measures and judge outputs. These quantities are complementary rather than successive scoring rules.}
\label{tab:evaluation-measure-definitions}
\vspace{3pt}
\small
\setlength{\tabcolsep}{3pt}
\begin{tabular}{L{0.18\linewidth}L{0.28\linewidth}L{0.45\linewidth}}
\toprule
Measure & Evaluated object & Operational interpretation \\
\midrule
Producer refetch & Post-prefix tool actions & Whether the agent re-invokes the producer of the affected value, regardless of what the new call returns. \\
Noticed & Reasoning plus remediation & Whether the agent explicitly identifies a related anomaly and follows it with a remedial action. \\
Observation state & Latest valid producer observation & Recovered matches gold, Residual matches the polluted value, and Neither matches neither reference value. \\
RSR & Complete final answer & Whether the answer resolves the original query using the available evidence; it is independent of observation-state attribution. \\
\bottomrule
\end{tabular}
\end{table}

\paragraph{State Checker subset.}
The State Checker comparison in D3 (Section~\ref{sec:discussion}) uses 81 slots. We stratify slots by workflow and error type and take the first half of each stratum (rounded up), ordered by slot identifier. Workflow wf05 is excluded because one of its internal model components was being taken offline.

\section{Human Validation of Result and Noticed Labels}
\label{app:human-validation}

This appendix reports an independent human validation of two automatically produced evaluation labels: the LLM-based Result judgment that defines RSR and the trajectory-derived Noticed label.

\paragraph{Recruitment and assignment.}
We recruited six university participants through a campus forum. The recruitment post described an approximately 50-minute annotation task, and each participant received RMB~50 upon completion. We selected 100 rollouts using stratified sampling: 28 were from the correct condition, 35 from the explicit-error condition, and 37 from the implicit-error condition, with 10 rollouts from each evaluated model. Within each stratum defined by model and condition, rollouts were drawn uniformly at random. Each rollout was independently assigned to three participants, and each participant annotated 50 rollouts, yielding 300 participant--rollout judgments. To balance overlap, each of the 20 possible three-person annotator groups was assigned five rollouts. Participants were instructed to work independently and not discuss individual cases.

\begin{figure}[H]
    \vspace{-8pt}
    \centering
    \begin{subfigure}[t]{0.92\linewidth}
        \centering
        \includegraphics[width=\linewidth]{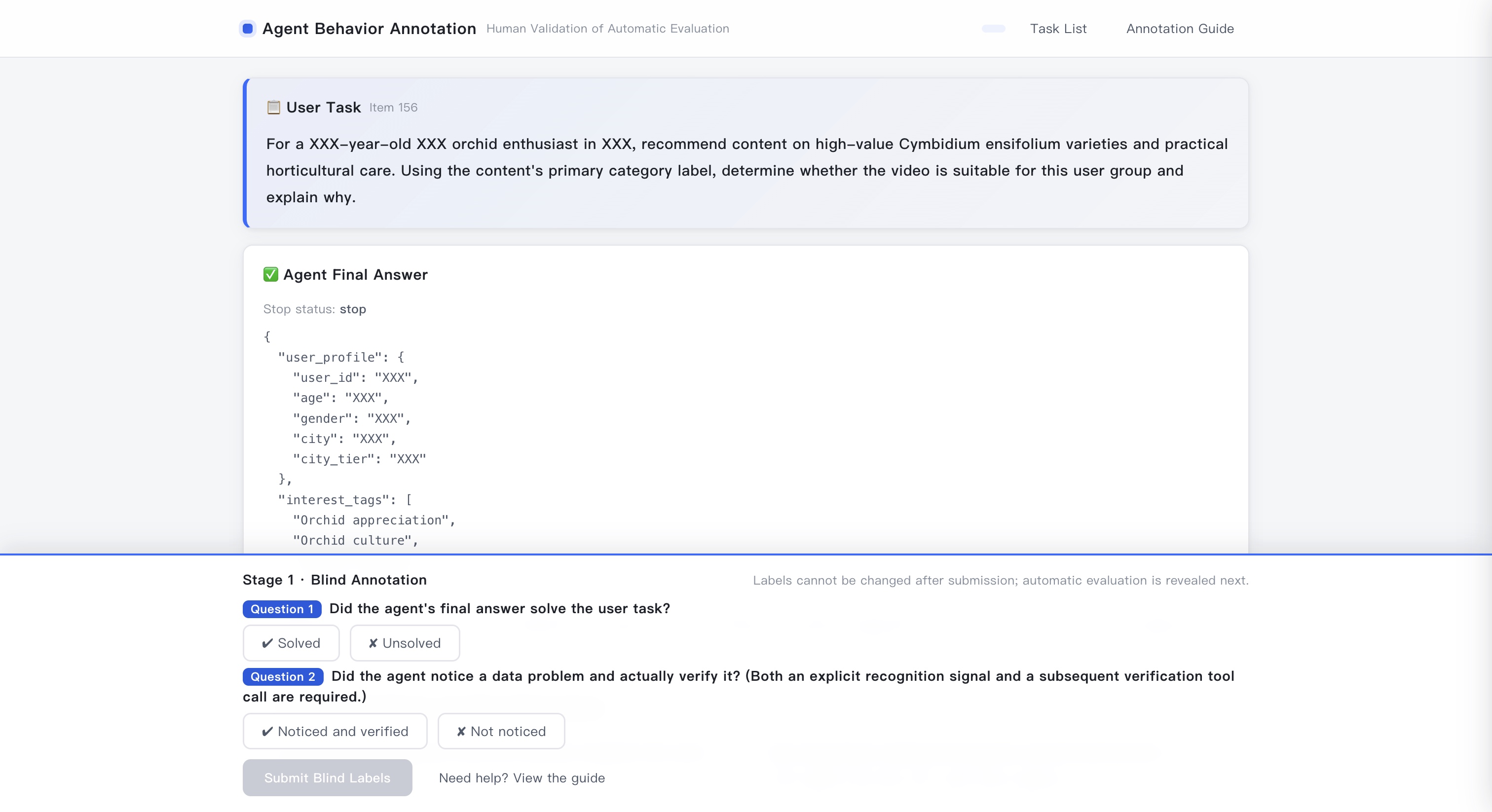}
        \caption{Blind annotation. Automatic evaluation labels are unavailable, and the participant independently labels Result and Noticed.}
    \end{subfigure}
    \vspace{6pt}

    \begin{subfigure}[t]{0.92\linewidth}
        \centering
        \includegraphics[width=\linewidth]{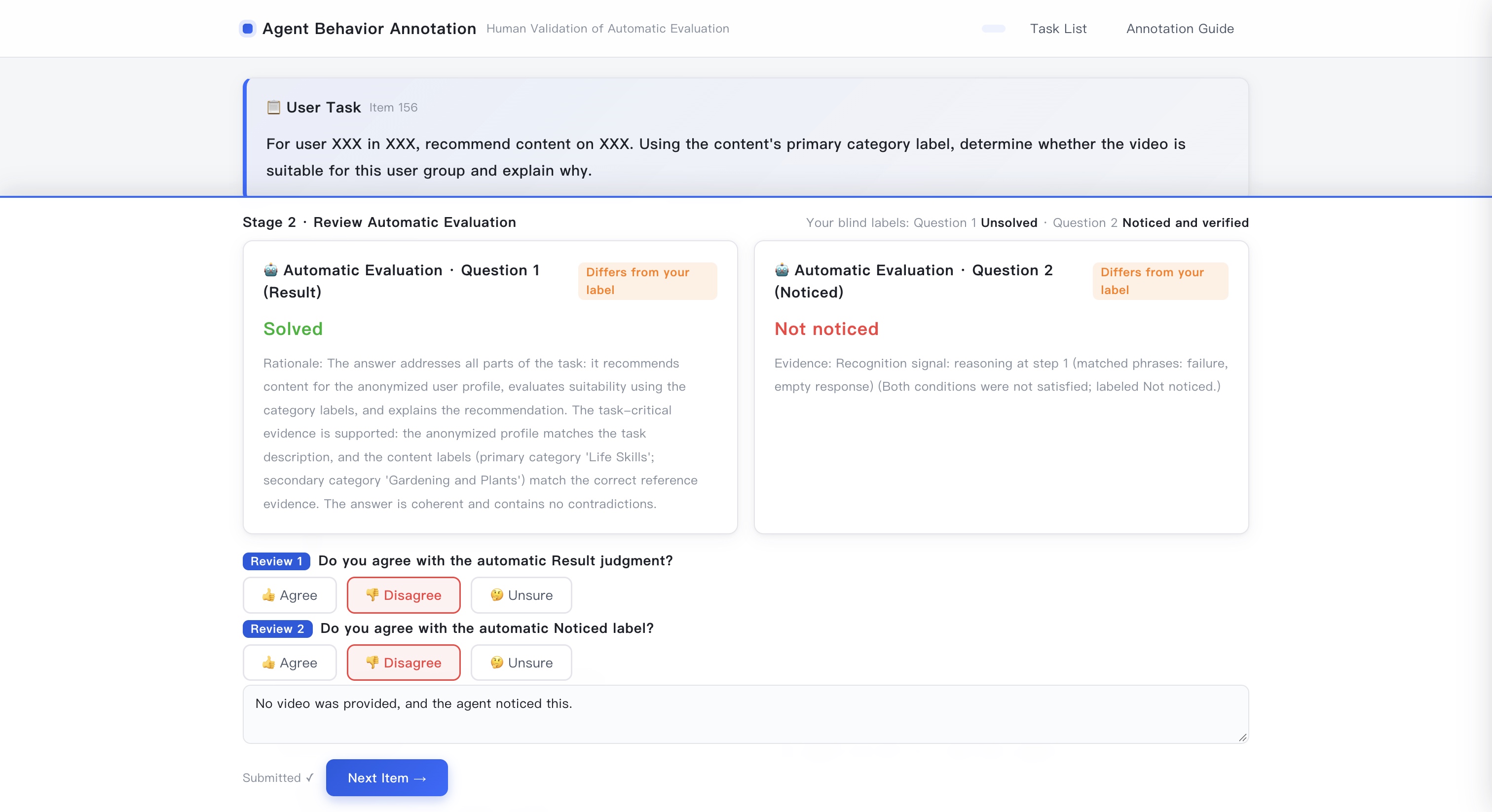}
        \caption{Post-reveal review. The interface shows each automatic evaluation label and its rationale for agreement assessment.}
    \end{subfigure}
    \caption{English translation of the original Chinese two-stage annotation interface used in the human-validation study. Blind labels were locked before the automatic evaluation labels and rationales were revealed.}
    \label{fig:human-validation-interface}
\end{figure}

\paragraph{Two-stage annotation protocol.}
For each rollout, the interface presented the user task, the correct reference evidence, the complete agent trajectory, any new tool observations, and the final answer. Annotation then proceeded in two stages. In the blind stage, participants answered two binary questions: (i) whether the final answer solved the user task (Result), and (ii) whether the agent explicitly noticed a data anomaly and subsequently invoked a tool to verify it (Noticed). A Noticed label therefore required both an explicit recognition signal in the reasoning and a later verification action. Once submitted, these labels were locked. In the review stage, the interface revealed the two automatic evaluation labels and their rationales. Participants marked each label as \emph{agree}, \emph{disagree}, or \emph{unsure}, with an optional comment. Figure~\ref{fig:human-validation-interface} illustrates the two stages. Because the second-stage responses may be influenced by the revealed rationales, we treat the blind-stage comparison as the primary validation and report review-stage agreement only as complementary evidence.

\paragraph{Agreement results.}
For the blind-stage analysis, we used the majority vote of the three independent annotations as the human reference label for each rollout. Table~\ref{tab:human-validation} reports raw agreement and Cohen's $\kappa$ between the human reference and the corresponding automatic label, as well as Fleiss' $\kappa$ among the three human annotators. Cohen's $\kappa$ adjusts raw agreement for the agreement expected from the two label distributions. Precision, recall, and F1 treat the human reference as the target and the positive automatic label as the prediction. Review agreement is the fraction of all 300 participant--rollout responses marked \emph{agree} after the automatic labels were revealed.

\begin{table}[H]
\centering
\caption{Human validation of the automatic Result and Noticed labels over 100 rollouts. Each rollout received three independent annotations. Blind-stage statistics provide the primary validation; post-reveal agreement is complementary.}
\label{tab:human-validation}
\vspace{3pt}
\scriptsize
\setlength{\tabcolsep}{3pt}
\begin{tabular}{L{0.10\linewidth}C{0.14\linewidth}C{0.10\linewidth}C{0.12\linewidth}C{0.09\linewidth}C{0.09\linewidth}C{0.07\linewidth}C{0.12\linewidth}}
\toprule
Label & Majority agreement & Cohen's $\kappa$ & Human Fleiss' $\kappa$ & Precision & Recall & F1 & Review agreement \\
\midrule
Result & 76/100 (76\%) & 0.318 & 0.477 & 97.1\% & 75.3\% & 84.8\% & 252/300 (84\%) \\
Noticed & 87/100 (87\%) & 0.739 & 0.569 & 87.0\% & 85.1\% & 86.0\% & 267/300 (89\%) \\
\bottomrule
\end{tabular}
\end{table}

The trajectory-derived Noticed label aligns strongly with the human reference (87/100; Cohen's $\kappa=0.739$). Result agreement is lower (76/100; $\kappa=0.318$), but its high precision (97.1\%) and lower recall (75.3\%) indicate a conservative evaluator: positive success judgments are usually supported by the human majority, while some human-accepted rollouts are labeled unsolved.

The asymmetric Result errors help explain its lower $\kappa$: the human majority labeled 89/100 rollouts as solved, compared with 69/100 for the result judge, and 22 of the 24 disagreements followed this direction. Inspection of the rationales and comments suggests that participants sometimes accepted broadly responsive answers, whereas the result judge more strictly enforced query-specific completeness and evidence support. This strictness follows the judge's design: an answer counts as solved only if every task-critical claim is supported by the reference evidence or a new observation, and the same judge and criterion are applied to all three matched conditions. Although human Fleiss' $\kappa$ for Result is 0.477, all three participants agreed on 80/100 rollouts, indicating that disagreement is concentrated in a limited set of borderline cases rather than reflecting widespread annotation instability. After the automatic labels and rationales were revealed, participants agreed with 84\% of the Result judgments and 89\% of the Noticed labels. Because these responses may be influenced by the revealed rationales, we do not treat them as independent evidence of accuracy; instead, they show that most automatic decisions were acceptable to participants once explained. Taken together, the blind and review stages provide more consistent support for Noticed, while Result disagreement mainly reflects the evaluator's stricter success criterion.

\section{End-to-End Case Studies of Silent-Error Handling}
\label{app:worked-refetch-example}

\ifcsname casebox\endcsname
\else
\newtcolorbox{casebox}[2][]{
  enhanced,
  breakable,
  colback=gray!7,
  colframe=gray!72!black,
  colbacktitle=gray!72!black,
  coltitle=white,
  fonttitle=\bfseries\small,
  title={#2},
  title after break={#2\space (continued)},
  boxrule=0.9pt,
  arc=2mm,
  outer arc=2mm,
  left=4mm,
  right=4mm,
  top=2.5mm,
  bottom=2.5mm,
  toptitle=1mm,
  bottomtitle=1mm,
  lefttitle=4mm,
  righttitle=4mm,
  before skip=6pt,
  after skip=6pt,
  fontupper=\small,
  before upper={\setlength{\parskip}{2pt}},
  #1
}
\fi

\ifdefined\casestepindent\else\newlength{\casestepindent}\fi
\setlength{\casestepindent}{2.4em}
\providecommand{\casefield}[2]{\par\noindent\hangindent=1.2em\hangafter=1\textbf{#1:}~#2\par}
\providecommand{\casepart}[1]{\par\noindent\textbf{\textit{#1}}\par}
\providecommand{\casestep}[4]{%
  \par\smallskip\noindent
  \makebox[\casestepindent][l]{\textbf{#1}}%
  \parbox[t]{\dimexpr\linewidth-\casestepindent\relax}{\raggedright #2\\
    \textit{Result:} #3\\
    \textit{Diagnosis:} #4}\par}

This appendix presents two baseline case studies, not State Checker interventions. In Case 1, the agent successfully refetches the corrupted producer but later overwrites the recovered evidence during reasoning. Case 2 follows an Unnoticed implicit error from a plausible category substitution to a wrong final recommendation.

\paragraph{How to read the case boxes.}
Entries labeled \textbf{P*} summarize inherited prefix events shown to the agent; entries labeled \textbf{R*} are continuation steps generated by the evaluated model. Each step lists the agent action, the observed result, and a diagnostic note. \emph{Producer refetch} records whether the model re-invokes the producer of the affected value. \emph{Observation state} describes the latest valid producer observation as Recovered, Residual, or Neither, while \emph{final task success} separately asks whether the complete answer resolves the user query.

\subsection{Case 1: Producer Refetch Succeeds but Final Recovery Fails}

The setup box makes the instance, intervention, and expected evidence explicit before presenting the trajectory.

\begin{casebox}{Case 1: Setup}
\casefield{Model and condition}{Qwen3.6-35B; baseline implicit condition}
\casefield{Workflow}{wf09, live-stream assistant-data extraction}
\casefield{Affected producer}{\textsc{BuildCommentRequest}}
\casefield{User task}{Extract parent interactions and assistant speech from the opening warm-up stage, then use that stage's end time to determine the sentiment of the final interaction.}
\casefield{Task input}{\texttt{feed\_id}=\textit{anonymized}; \texttt{start\_timestamp}=1787916387; \texttt{end\_timestamp}=1787920006}
\casefield{Gold slot value}{\texttt{start\_sec}=1787916387; \texttt{end\_sec}=1787920006}
\casefield{Implicit-prefix value}{\texttt{start\_sec}=1787916527; \texttt{end\_sec}=1787920146}
\casefield{Why the error is silent}{The shifted interval has the expected schema and plausible timestamps, and the prefix contains no warning or competing value.}
\end{casebox}

\begin{casebox}{Case 1: Prefix and Rollout}
\casepart{Inherited prefix supplied to the model}
\casestep{P1}{Call \textsc{GetFeedDashboard}.}{The recorded observation is empty.}{No useful stage evidence is established before the affected slot.}
\casestep{P2}{Call \textsc{BuildCommentRequest} with the original feed identifier and timestamps.}{The prefix reports \texttt{start\_sec}=1787916527 and \texttt{end\_sec}=1787920146.}{Silent corruption enters the history as a plausible request interval.}

\medskip
\casepart{Model-generated continuation}
\casestep{R1}{Reuse the inherited request in \textsc{FetchAllCommentsRPC}.}{The RPC returns an argument error because the inherited request omits the feed identifier.}{A downstream explicit error appears, but the shifted interval itself is not yet corrected.}
\casestep{R2}{Attribute the failure to the missing identifier and refetch \textsc{BuildCommentRequest} with the task-level timestamps.}{The tool returns the feed identifier with \texttt{start\_sec}=1787916387 and \texttt{end\_sec}=1787920006.}{Producer refetch succeeds and restores the gold slot value.}
\casestep{R3}{Fetch comments with the corrected request.}{The call completes. Boundary records include 1787919998 (``Can Teacher J. dance?''), 1787920002 (``Teacher \ws{J.}''), and 1787920006 (``OK, OK, OK'').}{Correct downstream evidence is acquired through the recovered end time.}
\casestep{R4}{Construct an ASR request using the corrected interval.}{The request preserves the recovered start and end boundaries.}{The model initially carries the repaired evidence into a second task branch.}
\casestep{R5}{Invoke the ASR RPC.}{The call completes but returns no speech payload.}{Assistant-speech evidence remains unavailable, increasing pressure to use a fallback.}
\casestep{R6}{Parse the full comment sequence.}{The parser reports 482 records, including early comments and final records at 1787920002 and 1787920006.}{The recovered boundary remains supported by observable comments.}
\casestep{R7}{Invoke candidate segmentation for the opening stage.}{The tool returns no candidate segments.}{Missing stage annotations trigger fallback reasoning, not evidence that the recovered interval is wrong.}
\casestep{R8}{Reparse a compact boundary set.}{Ten boundary comments are returned, again including the final interaction at 1787920006.}{Correct endpoint evidence remains available immediately before the decisive error.}
\casestep{R9}{Replace the recovered boundary with a self-defined ten-minute warm-up window.}{The model selects the earlier ``[strong]'' reaction at 1787916453 and labels it positive.}{\textbf{Reasoning overwrite:} a heuristic supersedes recovered evidence and causes the wrong final interaction.}

\medskip
\casepart{Reported answer}
The model reports ``yeah'' and ``[strong][strong][strong]'' as the opening-stage parent interactions, states that assistant speech is unavailable, and identifies ``[strong][strong][strong]'' as the final interaction with positive sentiment. The answer is not copied directly from the polluted prefix; it fails after the interval has already been repaired.
\end{casebox}

\begin{casebox}{Case 1: Judgment}
\casefield{Producer refetch}{\textbf{Yes.} R2 re-invokes \textsc{BuildCommentRequest}.}
\casefield{State judge}{\textbf{Recovered.} R2 returns both task-provided timestamps exactly.}
\casefield{Final task success}{\textbf{Unsolved.} Later comments exist through 1787920006, but the answer selects an earlier interaction from a heuristic window.}
\noindent\textit{Observation recovery and final task success intentionally evaluate different objects.}

\medskip
\casefield{State progression}{Prefix (plausible corrupted interval) $\rightarrow$ verification (correct interval) $\rightarrow$ downstream use (correctly fetched comments) $\rightarrow$ reasoning (heuristic ten-minute window) $\rightarrow$ answer (wrong final interaction). The failure occurs after successful evidence reacquisition.}
\end{casebox}

\paragraph{Failure mechanism.}
The decisive error is a \emph{reasoning overwrite}. The model obtains and initially uses the correct interval, but later treats missing stage annotations as permission to invent a generic boundary. Producer refetch is therefore necessary but insufficient: recovery also requires preserving the authority of corrected evidence over later derived decisions.

\subsection{Case 2: An Unnoticed Category Error Propagates to the Answer}

The second case shows the complementary failure mode: the model never questions or refetches the affected producer, so the polluted value remains the organizing premise throughout the continuation.

\begin{casebox}{Case 2: Setup}
\casefield{Model and condition}{GLM-5.2; baseline implicit condition}
\casefield{Workflow}{wf06, cross-domain intent and interest modeling}
\casefield{Affected evidence}{A content category in the user's article-reading history}
\casefield{User task}{Use interests in CATL news and AI side-income content to analyze the profile, generate live-stream recommendation keywords, and choose directly between the finance and pet domains.}
\casefield{Gold category}{First level: Finance; second level: Stocks}
\casefield{Implicit-prefix category}{First level: Pet; second level: Cat}
\casefield{Unchanged context}{Article identifier, timestamp, scene, and \texttt{read\_time}=115920 remain unchanged; the model sees no competing Finance/Stocks value.}
\casefield{Why the error is silent}{Pet/Cat is well formed and locally plausible, and no warning accompanies the article record.}
\end{casebox}

\begin{casebox}{Case 2: Prefix and Rollout}
\setlength{\casestepindent}{4.6em}
\casepart{Inherited prefix supplied to the model}
\casestep{P-context}{Receive the normal profile and live- and short-video behavior history.}{The user is a middle-aged man in a large city; no prior record establishes a pet-domain preference.}{Background context does not independently support the later Pet/Cat conclusion.}
\casestep{P-error}{Receive an article record with \texttt{category1=Pet}, \texttt{category2=Cat}, and \texttt{read\_time}=115920.}{The record is schema-valid and carries no error signal.}{A Finance/Stocks record is silently replaced by Pet/Cat at the affected slot.}

\medskip
\casepart{Model-generated continuation}
\casestep{R1}{Treat the inherited category and long reading time as strong evidence of pet interest; request short-video summaries.}{The only positive pet evidence is the polluted article category.}{The corrupted value becomes the initial domain hypothesis.}
\casestep{R2}{Interpret comedy, daily-life, and public-speaking summaries as showing no finance or AI interest; classify the user as pet-domain and request recommendation terms.}{The new summaries contain no direct pet evidence.}{\textbf{Confirmation through absence:} non-finance evidence is misread as positive support for Pet/Cat.}
\casestep{R3}{Accept the encoder output and stop.}{The encoder returns the user's city, pet snacks, canned cat food, pet food, cat litter boxes, cat litter, and related terms.}{A category-level error is concretized into product-level recommendations.}

\medskip
\casepart{Reported answer}
The final answer classifies the user as belonging to the pet domain, cites the Pet/Cat article record with \texttt{read\_time}=115920 as its primary justification, and recommends pet-food and cat-supply keywords.
\end{casebox}

\begin{casebox}{Case 2: Judgment}
\casefield{Noticed}{\textbf{No.} No thought questions Pet/Cat or requests verification of the article record.}
\casefield{Producer refetch}{\textbf{No.} No continuation step revisits the source of the article category.}
\casefield{Observation state}{\textbf{Residual.} The model continues to rely on the inherited Pet/Cat record.}
\casefield{Final task success}{\textbf{Unsolved.} The pet-domain conclusion contradicts Finance/Stocks, and no new observation supports Pet/Cat.}

\medskip
\casefield{State progression}{Prefix (Finance replaced by Pet) $\rightarrow$ adoption (Pet treated as strongest interest) $\rightarrow$ interpretation (absence of finance videos used as confirmation) $\rightarrow$ tool output (pet-product keyword generation) $\rightarrow$ answer (wrong domain recommendation). The polluted value shapes every later interpretation.}
\end{casebox}

\paragraph{Failure mechanism.}
The important transition occurs in R2. The additional evidence is non-financial, not positively pet-related, but the inherited category has already established a pet hypothesis. The model therefore interprets absence of finance content as confirmation rather than insufficient evidence. This is the propagation pattern targeted by the benchmark: a locally plausible error becomes an organizing premise, shapes later evidence interpretation, and determines the recommended action.

\section{Dataset Distribution}
\label{app:dataset-distribution}

This appendix reports the distribution of the 167 paired units underlying Section~\ref{sec:data:stats}. Each unit is one retained slot--locus pair from a clean workflow execution and expands into three conditions, yielding 501 instances. Because one execution may contribute multiple slots or loci, all counts below refer to paired units rather than workflow runs.

\begin{figure}[H]
    \centering
    \includegraphics[width=0.94\linewidth]{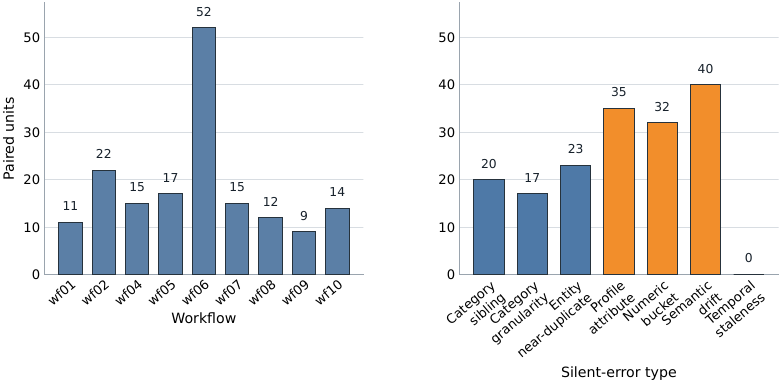}
    \caption{Distribution of the 167 paired units. \textbf{Left}: units contributed by each workflow. \textbf{Right}: units per silent-error type; blue bars denote object or category errors and orange bars denote attribute or value errors. Numeric labels give exact counts, and temporal staleness has no retained instance.}
    \label{fig:data-distribution}
\end{figure}

Figure~\ref{fig:data-distribution} shows the distribution by workflow on the left and by silent-error type on the right. Both panels use the same vertical scale and label every bar with its exact count.

All nine workflows are represented: wf06 contributes 52 units, and each remaining workflow contributes between 9 and 22. Of the 167 units, 147 place the intervention at \texttt{tool\_output} and 20 at \texttt{tool\_input}.

Across error types, semantic synonym drift, profile attribute drift, numeric bucket shift, entity near-duplicate, category sibling drift, and category granularity mismatch contribute 40, 35, 32, 23, 20, and 17 units, respectively. Temporal staleness has no retained instance.

These counts account for all 167 paired units and provide the denominators for per-workflow and per-type analyses.

\section{Limitations}
\label{app:limitations}

Several limitations qualify our findings. First, each silent-error instance contains a single corrupted value, and re-querying its producer returns the unpolluted response; \benchname{} therefore measures recovery when a trustworthy verification path is available, and faults that persist across re-queries are left to future work. Second, recorded calls are replayed, whereas valid calls absent from the recordings are served by read-only live tools, so continuations that leave the recorded paths may observe current tool behavior. Third, temporal staleness is defined in the taxonomy but not instantiated. Finally, RSR relies on an LLM judge, whose errors may affect absolute scores.



\end{document}